%% file: main.tex
\documentclass[preprint,12pt]{elsarticle}

\usepackage{amssymb}
\input{math_commands.tex}

\usepackage{hyperref}
\usepackage{url}
\usepackage[table,xcdraw]{xcolor}
\usepackage{graphicx}
\usepackage{multirow}
\usepackage{svg}
\usepackage{enumitem}
\usepackage{float}

\journal{Computer Vision and Image Understanding}

\makeatletter

\def\myfooterfont#1{\gdef\@myfooterfont{#1}}
\myfooterfont{\footnotesize\itshape}

\gdef\@elsarticlemyfooter{%
Accepted author manuscript of the paper published in \emph{Computer Vision and Image Understanding} (2026). The final version is available at \url{https://doi.org/10.1016/j.cviu.2026.104698}.%
}

\def\ps@pprintTitle{%
  \let\@oddhead\@empty
  \let\@evenhead\@empty
  \def\@oddfoot{%
    \hbox to \textwidth{%
      \hfil
      \parbox[b]{0.9\textwidth}{%
        \@myfooterfont
        \@elsarticlemyfooter
      }%
      \hfil
    }%
  }%
  \let\@evenfoot\@oddfoot
}

\def\@seccntDot{.}
\def\@seccntformat#1{\csname the#1\endcsname\@seccntDot\hskip 0.5em}

\makeatother

\begin{document}

\begin{frontmatter}



\title{Escaping the big data paradigm in self-supervised representation learning}


\author[inst1]{Carlos Vélez-García\corref{cor1}}

\affiliation[inst1]{organization={Automation \& Robotics},%Department and Organization
            addressline={INESCOP}, 
            city={Elda},
            postcode={00000}, 
            state={Alicante},
            country={Spain}}

\author[inst2]{Miguel Cazorla}
\affiliation[inst2]{organization={Institute for Computer Science},%Department and Organization
            addressline={University of Alicante}, 
            city={San Vicente del Raspeig},
            postcode={E03690}, 
            state={Alicante},
            country={Spain}}
\author[inst3]{Jorge Pomares}
\affiliation[inst3]{organization={DFISTS},%Department and Organization
            addressline={University of Alicante}, 
            city={San Vicente del Raspeig},
            postcode={E03690}, 
            state={Alicante},
            country={Spain}}

\cortext[cor1]{Corresponding author: cvg25@alu.ua.es }
\begin{abstract}
The reliance on large-scale datasets and extensive computational resources has become a significant barrier to advancing representation learning from images, particularly in domains where data is scarce or expensive to obtain. In this paper, we address the critical question: \textit{Can we escape the big data paradigm in self-supervised representation learning from images?} We introduce \textbf{SCOTT} (\textbf{S}parse \textbf{Co}nvolutional \textbf{T}okenizer for \textbf{T}ransformers), a shallow tokenization architecture that is compatible with Masked Image Modeling (MIM) tasks. SCOTT injects convolutional inductive biases into Vision Transformers (ViTs), enhancing their efficacy in small-scale data regimens. Alongside, we propose to train on a Joint-Embedding Predictive Architecture within a MIM framework (\textbf{MIM-JEPA}), operating in latent representation space to capture more semantic features. Our approach enables ViTs to be trained from scratch on datasets orders of magnitude smaller than traditionally required —without relying on massive external datasets for pretraining. We validate our method on three small-size, standard-resolution, fine-grained datasets: Oxford Flowers-102, Oxford IIIT Pets-37, and ImageNet-100. Despite the challenges of limited data and high intra-class similarity of these datasets, our frozen SCOTT models pretrained with MIM-JEPA significantly outperform fully supervised methods and achieve competitive results with state-of-the-art approaches that rely on large-scale pretraining, complex image augmentations and bigger model sizes. By demonstrating that robust off-the-shelf representations can be learned with limited data, compute, and model sizes, our work paves the way for computer applications in resource constrained environments such as medical imaging or robotics. 
\end{abstract}


%\item SCOTT: MIM-compatible convolutional tokenizer injecting CNN biases into ViTs via a sparse architecture.
%\item MIM-JEPA: JEPA on masked image modeling for sample-efficient visual representation learning.
%\item Data-efficient ViTs: SCOTT+MIM-JEPA outperforms prior methods in limited-data regimes.
%\item Low-resource SSL: matches large foundational SSL models under modest compute budgets.

\begin{keyword}
Data-efficient representation learning \sep self-supervised learning \sep sparse image tokenization.
\end{keyword}

\end{frontmatter}


\section{Introduction}
Escaping the big data paradigm in self-supervised learning from images is crucial for the future of computer vision (CV). Representation learning, described in \citep{bengio_representation_2014} as “learning representations of the data that make it easier to extract useful information when building classifiers or other predictors”, becomes particularly relevant when training data is scarce as it would enable efficient learning for downstream tasks. Traditionally, transfer learning has been the dominant approach, where convolutional neural networks (CNNs) \citep{lecun_backpropagation_1989} are pretrained on large-scale labeled datasets like ImageNet \citep{deng_imagenet} and then fine-tuned on specific tasks. However, this approach has two major constraints: the reliance on vast labeled datasets for pretraining and the domain-specific brittleness of the learned features \citep{jain_data-based_2023}. These limitations make it impractical in fields like medical imaging or industrial applications, where data collection requires domain-expertise and is both time-consuming and expensive \citep{huang_self-supervised_2023}.

In recent years, self-supervised learning (SSL) has emerged as a promising alternative, motivated by the success of methods such as BERT \citep{Devlin_BERTPO} and GPT \citep{radford2019language} in natural language processing (NLP). The core idea behind SSL is to devise a task that provides a supervisory signal from the data itself without explicit human annotation, allowing models to learn meaningful representations in a label-free environment \citep{caron2021emerging}. However, self-supervised learning success in both NLP and CV must largely be attributed to the advent of the Transformer architecture \citep{vaswani2017attention}, which leverages self-attention mechanisms to capture long-range dependencies in data in a highly parallel and scalable manner. The Vision Transformer (ViT) \citep{dosovitskiy_image_2021} marked the first significant attempt to apply a purely transformer architecture to visual tasks, but its success hinges on access to extremely large datasets (14M-300M images) \citep{deng_imagenet, sun_revisiting_2017, asano_pass_2021}. As ViT authors noted, Transformers lack certain inductive biases inherent to CNNs -such as translation equivariance and locality- which makes them less effective when trained on limited data \citep{dosovitskiy_image_2021}.

Over the past few years, this combination of label-free training methods with ViT has led to a “resource-hungry” training paradigm, with most research efforts pushing the state of the art in the natural image domain through scaling to even larger models and dataset sizes. Unfortunately, this trend limits major contributions from researchers with limited compute and data budgets and poses significant challenges in specialized fields where domain-specific data is difficult to acquire. Therefore, escaping the big data paradigm is crucial for advancing computer vision applications in fields beyond natural images. By reducing the dependency on large datasets, we could make advancements in this field more accessible and impactful across a wider range of applications \citep{huang_self-supervised_2023}.

\begin{figure}[t]
    \centering
    \includegraphics[width=\textwidth]{"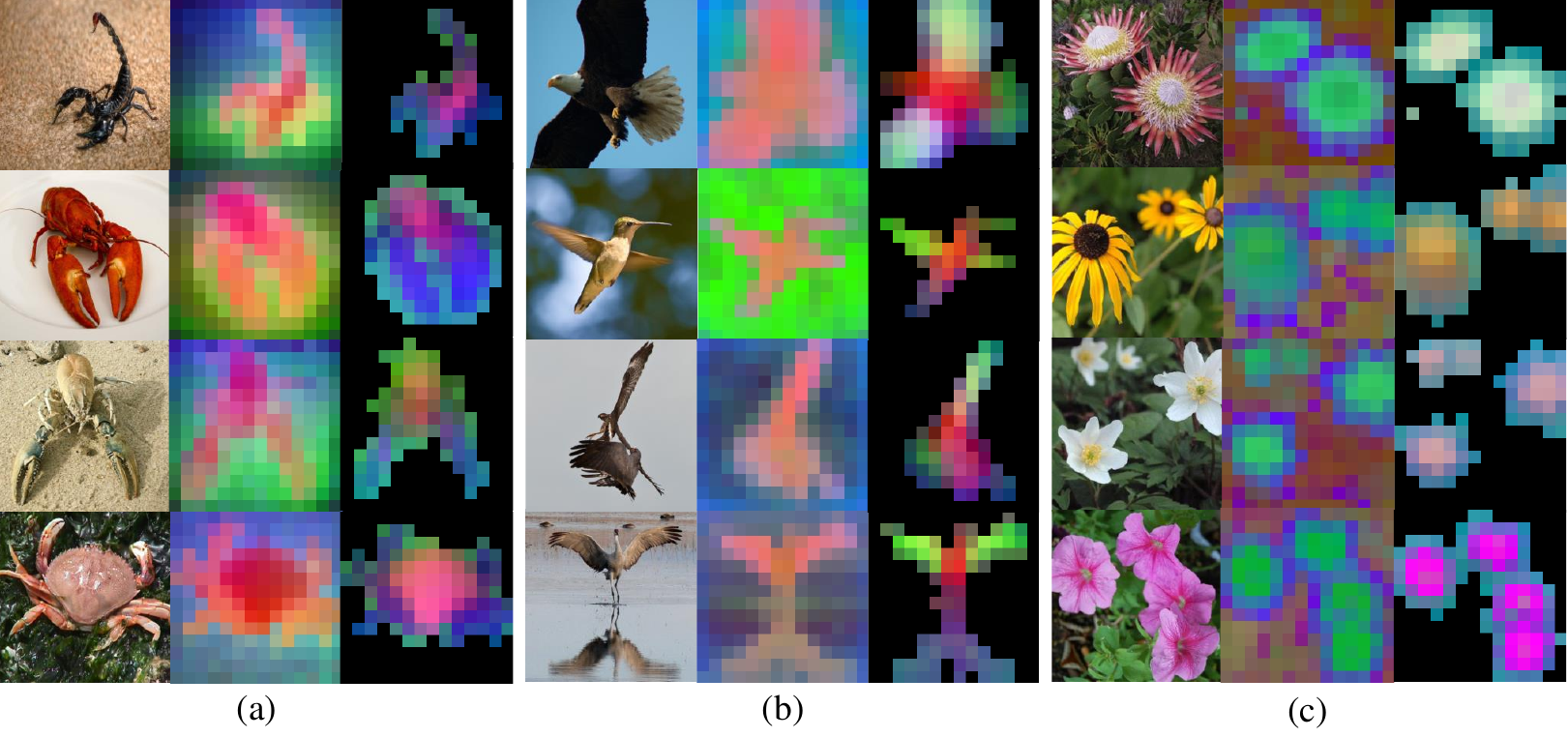"}
    \caption{\textbf{Matching different semantic parts across categories and poses}. We show the first 3 components of a PCA computed among the token embeddings of images from the same column (a, b, and c). The background is removed by thresholding the first component. Notably, semantically similar parts are matched by color despite belonging to different object classes and poses. For instance: in (a) animal claws are purple and torso pink, in (b) wings are green and torso red. Interestingly, once background is removed in (c), different flower disks are matched to different colors. }
    \label{fig:pca_many}
\end{figure}

Thus, a pressing question arises: \textbf{Can we escape the big data paradigm in self-supervised representation learning from images?}

We interpret ``escaping'' as learning transferable representations from scratch using only in-domain unlabeled data and moderate compute, without relying on large-scale external pretraining. Recent work suggests that masked predictive learning can yield strong visual representations without supervision, and that invariances to occlusion or masking can be beneficial in few-sample settings \citep{kong2023understanding}. However, in the low-resource regime considered here---training Vision Transformers (ViTs) from scratch on only a few thousand images---we observe an additional practical limitation: under aggressive masking, standard patchify tokenization provides weak locality priors and discards boundary continuity cues early in the network, which can make optimization less stable and reduce the quality of the learned features. This issue is amplified in the absence of large-scale pretraining, where architectural inductive biases and faithful propagation of the masking signal become more important.

In this work, we take a step towards addressing this challenge by introducing two key contributions: the Sparse Convolutional Tokenizer for Transformers (SCOTT) and a Joint-Embedding Predictive Architecture (JEPA) for vision instantiated in a Masked Image Modeling (MIM) framework, which we refer to as (MIM-JEPA). Rather than proposing a new objective alone, we focus on architectural compatibility with masked predictive learning as a prerequisite for sample-efficient representation learning. SCOTT is a tokenization architecture that replaces the original patch-based embedding of ViTs, and not only incorporates the inductive biases of CNNs to allow ViT to operate effectively in small-scale data regimes, but also its sparsity mitigates issues like information leakage and mask vanishing, which have previously hindered the application of MIM strategies in CNN-based tokenizers for transformers. Moreover, in contrast to generative MIM methods that predict missing information in pixel/token space, the JEPA objective is in abstract representation space where unnecessary pixel-level details are potentially eliminated, leading the model to produce more semantic features. This capability is demonstrated in Figures \ref{fig:pca_many} and \ref{fig:pca_duck}, where we apply a principal component analysis (PCA) on the patch features produced by our method, revealing meaningful semantic structures.

To prove our method’s potential to unlock deep learning for the long tail of vision tasks without expensive labelled datasets, we constrain our research to three small-size, standard-resolution, fine-grained datasets. Specifically, we focus on two popular computer vision datasets from the VTAB benchmark \citep{zhai_large-scale_2020}: Oxford Flowers-102 \citep{nilsback_automated_2008} and Oxford IIIT Pets-37 \citep{parkhi_cats_2012}; the third one is ImageNet-100 \citep{deng_imagenet}, a subset of the well-studied ImageNet with 100 different classes of animals. Apart from the small data available for training, with roughly 20 samples per class in Flowers-102, these datasets present a significant challenge due to their high intra-class similarity, making them ideal for testing the limits of self-supervised learning without large datasets. It is worth noting that unlike previous works \citep{dosovitskiy_image_2021, bao2021beit, assran2023self, oquab_dinov2_2024, zhou2021ibot}, \citep{steiner_how_2022}, that rely on pretraining on massive external datasets for learning to see \citep{steiner_how_2022}, our method is trained entirely from scratch using only the images, without labels, of the target dataset. Due to computational constraints, we were unable to evaluate on ImageNet-1K scale and beyond and leave this scalability assessment for future work.

Our contributions are motivated by the observation that, in small-data regimes, masked predictive objectives can underperform not only due to the choice of objective, but also due to architectural mismatches that weaken local signal propagation under heavy masking. We therefore propose an architecture that preserves locality and stable feature extraction while remaining compatible with masked training, and we validate that this enables strong performance without relying on large-scale external pretraining.

In summary, our contributions are as follows:
\begin{itemize}
    \item We propose SCOTT, a \textbf{S}parse \textbf{Co}nvolutional \textbf{T}okenizer for \textbf{T}ransformers that injects CNN-like inductive biases into ViTs and is compatible with masked image modeling (MIM) training due to its sparse architecture.
    \item We demonstrate that instantiating a Joint-Embedding Predictive Architecture (JEPA) on a MIM task —referred to as MIM-JEPA— yields sample-efficient representations that boost performance on fine-grained visual recognition tasks.
    \item We show that combining SCOTT with the MIM-JEPA framework enables Vision Transformers to perform effectively in small-scale environments, significantly outperforming previous methods and drastically reducing reliance on large datasets.
    \item We ensure our method remains accessible to researchers with modest computational resources, making state-of-the-art self-supervised learning more inclusive and adaptable.
\end{itemize}

Through this work, we aim to advance self-supervised learning in computer vision by making it more accessible and practical for a broader spectrum of applications, particularly in domains where large-scale datasets are not feasible. We present an efficient model with few parameters, that can be quickly and effectively trained on smaller platforms while still maintaining state-of-the-art results.

\section{Related Works}
\subsection{Injecting ViT with Convolutional Priors}
Vision Transformer (ViT) reliance on large datasets stems from the lack of inductive biases inherent to convolutional neural networks (CNNs) \citep{dosovitskiy_image_2021}. CNNs, inspired by the hierarchical processing of the mammalian visual cortex \citep{hubel_receptive_1959, fukushima_neocognitron_1988}, provide important priors for learning spatial relationships in visual data. Recognizing this limitation, numerous studies have previously explored incorporating convolutional priors into ViT architectures \citep{wu_cvt_2021, chen_visformer_2021, yuan_incorporating_2021, graham_levit_2021}. Early attempts, such as the “hybrid ViT” \citep{dosovitskiy_image_2021}, fed a ResNet \citep{he2016deep} feature map into a transformer encoder, showing a slight performance advantage over ViT at smaller computational budgets. However, later studies \citep{xiao2021early} revealed that excessive convolutional layers could diminish the generalization power of ViTs, suggesting that a shallow convolutional stem might strike the right balance between CNN-like inductive biases and the representational power of transformers.

The Compact Convolutional Transformer (CCT) \citep{hassani_escaping_2022} followed this principle, introducing a convolutional tokenizer tailored for supervised learning on datasets significantly smaller than ImageNet. While CCT demonstrates strong performance on supervised contexts, its reliance on standard convolutions makes it fundamentally incompatible with masked image modeling self-supervised tasks.

Building upon this foundation, our work pushes the idea further by leveraging sparse convolutions \citep{liu_sparse_2015} - a design choice introduced by the Spark framework \citep{tian_designing_2023}, to enable BERT pre-training on CNN architectures - to enhance tokenization specifically for self-supervised learning. Unlike Spark, which employs a fully convolutional encoder-decoder architecture reminiscent of U-Net and adopts a generative BERT-style training framework, our proposed sparse convolutional tokenizer architecture is tailored to overcome critical limitations such as information leakage and mask vanishing that have previously hampered the application of traditional convolution-based tokenizers for vision transformers in MIM tasks.

\subsection{Masked Predictive Representation Learning}
Masked Image Modeling (MIM), first introduced in BEiT \citep{bao2021beit}, draws inspiration from the success of BERT in NLP \citep{Devlin_BERTPO}. In this approach, an image is divided in non-overlapping patches and a subset of these patches is masked out. The model is tasked with reconstructing the masked regions, which encourages learning meaningful representations of visual features, akin to how BERT learns semantic dependencies in text. Since its introduction, MIM has been extended with different reconstruction targets, such as raw pixels \citep{he2022masked, xie2020unsupervised, xie_simmim_2022},  or patch-level tokens via learned tokenizers \citep{bao2021beit, peng_beitv2_2022}. While effective at scale, these approaches often produce low-level features that lack the semantic abstraction needed for fine-grained tasks. To address this, invariance-based methods (e.g., DINO, MoCo, SimCLR) \citep{zhou2021ibot, oquab_dinov2_2024} have been combined with MIM objectives, yielding more semantic features \citep{kong2023understanding}. However, these methods rely heavily on complex, hand-crafted augmentations, which can introduce biases detrimental to some downstream tasks \citep{assran2023self} and may not generalize well to other scientific domains \citep{huang_self-supervised_2023}. Unlike invariance-based methods such as DINO, SimSiam, or MoCo, which rely on multi-view augmentations and large memory/computation budgets, our approach is explicitly designed for practitioners in constrained-resource settings, where such methods are impractical.

Our work is directly inspired by I-JEPA \citep{assran2023self}, which takes this concept further by predicting masked abstract targets in the representation space produced by a momentum-based target-encoder ViT network. However, as previous MIM-based approaches, I-JEPA is evaluated exclusively on large-scale pretraining to achieve competitive results with other foundational vision models. Building on these ideas, we integrate our Sparse Convolutional Tokenizer for Transformers (SCOTT) within the ViT architecture of a JEPA framework based on MIM, which we refer to as MIM-JEPA. 

Our proposed MIM-JEPA differs from prior work in two respects: (i) it integrates SCOTT to process both visible and masked tokens, resolving the incompatibility of CNN stems with MIM; and (ii) it is designed to train from scratch on small datasets, showing that in domain state-of-the-art performance is achievable without large-scale pretraining.

\textbf{Occlusion-invariant masked modeling.} Closely related to our objective, \citep{kong2023understanding} interpret masked image modeling through the lens of learning occlusion-invariant features and propose an occlusion-invariant training objective (C-MAE) that can improve transfer in low-label settings. Since this line of work directly targets the key question posed in this paper—escaping the big-data paradigm—we implement both MAE \citep{he2022masked} and C-MAE \citep{kong2023understanding} as baselines under the same compute and evaluation constraints as our method (Sec. \ref{sec:ablations}).

\section{Method}
\begin{figure}[t]
    \centering
    \includegraphics[width=\textwidth]{"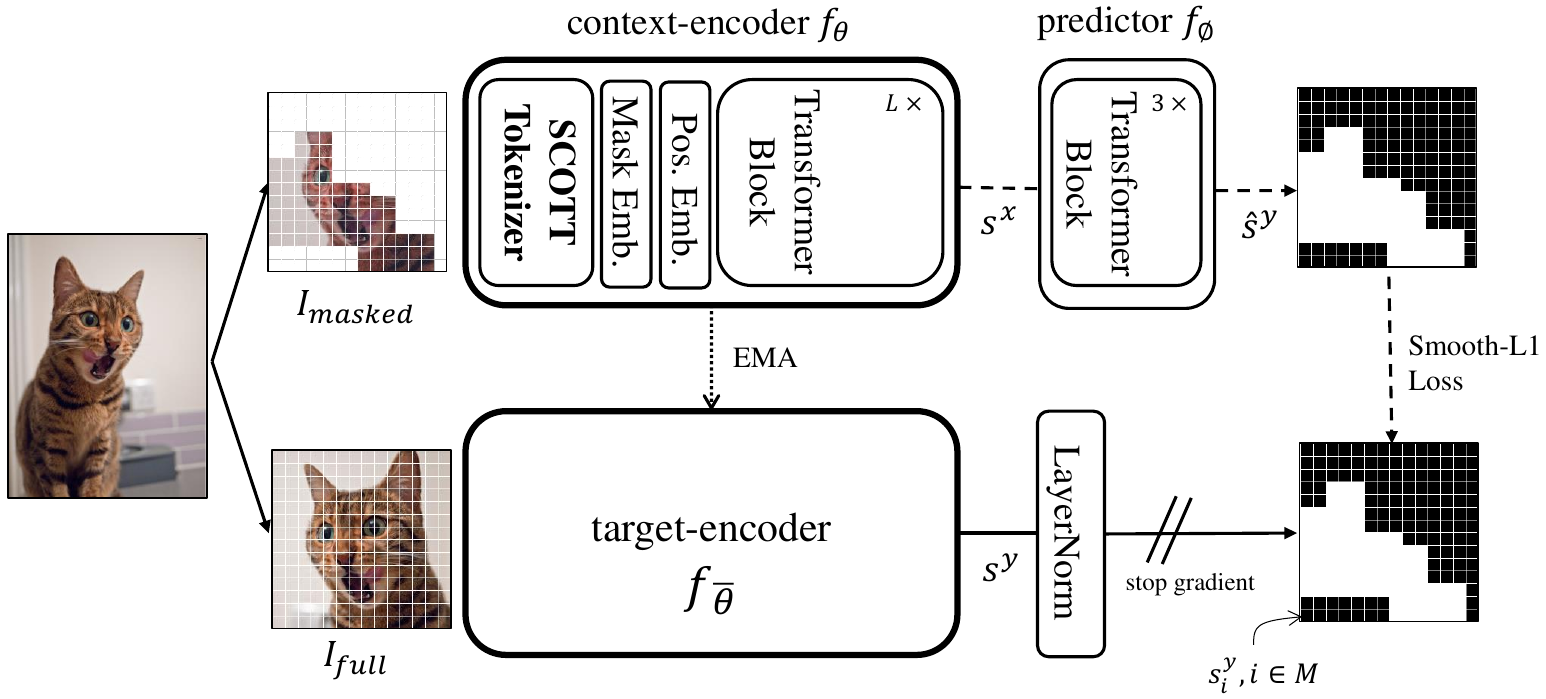"}
    \caption{\textbf{MIM-JEPA}. An image $I_{full}$ is processed by the target-encoder $f_{\bar{\theta}}$ to produce a latent patch-level representation $s^y$, whose masked patches $M$ are used as targets; The context image $I_{masked}$, generated from the complement of $M$, is input to the context-encoder $f_{\theta}$ to produce $s^x$. The predictor $f_\phi$ is fed with $s^x$ to predict the missing content $\hat s^y$. The Smooth-L1 loss is computed only on the (black) masked patches in latent space to update the context-encoder and predictor weights (dashed line), while the target encoder's weights are updated via an exponential moving average (EMA) of the context-encoder (dotted line).}
    \label{fig:method}
\end{figure}

To provide empirical evidence that ViTs can be effectively trained from scratch on small datasets, we propose to harness the full power of self-supervised learning for learning representations. To this end, we design a Joint-Embedding Predictive Architecture instantiated through a MIM task, referred to as  MIM-JEPA, and illustrated in Figure \ref{fig:method}.

The overall training objective is as follows: given a masked image as input to a context-encoder, a predictor is tasked with learning the latent representations of the masked blocks produced by a target-encoder that processes the full image. Furthermore, to address the suboptimal optimizability of ViTs caused primarily by the \textit{patchify} stem (i.e., tokenizer), we propose to replace it by a Sparse Convolutional Tokenizer for Transformers (SCOTT). This tokenizer is compatible with MIM objectives and injects convolutional priors into ViTs, offering superior data efficiency and performance.

\subsection{Sparse Convolutional Tokenizer for Transformers (SCOTT)}
\label{scott}
A standard transformer \citep{vaswani2017attention} takes as input a sequence of vectors, called tokens. However, there is a fundamental difference between the signal space of NLP and the signal space of computer vision, given that language data is discrete and structured (i.e., words), whereas image content is high dimensional, continuous, and unstructured (i.e., pixel values) \citep{ozbulak_know_2023}.

Image tokenization in standard ViTs is performed by a patch and embed layer which subdivides an image into non-overlapping square patches so that a transformer can accept visual data. Formally, the image $x{\in}R^{H\times W\times C}$ is reshaped into $N=HW/P^2$ patches $x^p{\in}R^{H\times (P^2C)}$, where $C$ is the number of channels, $H, W$ is the input image resolution, and $(P, P)$ is the resolution of each patch. The image patches $\{x_i^p\}_{i=1}^N$ are then linearly projected into patch embeddings $\{e_i^p\}_{i=1}^N$ each with dimension $d$. This is equivalent to a convolutional layer with $d$ filters, and $P \times P$ stride and kernel size. Among other limitations, this simple patch and embedding method eliminates the boundary-level information present in different patches.

Specifically in our experiments, we split each $224\times 224$ image into a $14\times 14$ grid of patch embeddings, where each embedding corresponds to a $16\times 16$ image patch.

In order to inject some inductive biases into the transformer architecture, we propose to replace the patch and embedding in ViT by a shallow convolutional stem. This stem follows conventional design, which consists of 2 consecutive blocks of: convolution, ReLU activation, and a max blur pool layer \citep{zhang2019making} (see \ref{app:scott-arch} for details). The output of the convolutional stem proposed produces $\{e_i^p\}_{i=1}^N$, a $14\times 14$ feature map each with dimension $d$ matching the number of inputs to the transformer created by the standard patch and embedding method. 

However, introducing a CNN tokenizer conflicts with the patch-wise masking strategy because one cannot eliminate pixel information from masked patches - to avoid trivial solutions - as ViTs do by removing or replacing them with a mask token. Setting masked patches to zero in CNNs has drawbacks: (i) it disturbs the pixel value distribution; (ii) masked patterns vanish after several convolutional layers; (iii) computations on masked regions are unnecessary. To overcome this, we gather masked patches into a sparse image and employ sparse layers that compute only when the kernel center covers a non-empty element (see "submanifold sparse convolution" in \citep{graham_submanifold_2017}). Since dense images are special cases of sparse images without holes, sparse layers naturally reduce to standard ones when masking isn't applied.

\textbf{SCOTT enabled Vision Transformer.} Following ViT \citep{dosovitskiy_image_2021}, our backbone network is a standard Transformer \citep{vaswani2017attention} to ensure a fair comparison between our results and previous works in terms of network architecture. Specifically, our ViT can be decomposed in parts: SCOTT for image tokenization, fixed sinusoidal Positional Embedding, a Mask Token, and $L$ consecutive Transformer Encoder blocks. Since our method learns representations without labels, we do not use a class token nor a classification head present in the standard ViT. The features used in downstream tasks are the model’s frozen output. We use similar notation in ViT for SCOTT enabled variants: for instance, SCOTT-7/16 is a vision transformer that has a SCOTT tokenizer with a patch size of 16 and 7 transformer encoder blocks.

\subsection{Learning Image Representations (MIM-JEPA)}
\label{subsec:mim-jepa}
We first formally describe Masked Image Modeling (MIM) which lays the foundation for then proposing the MIM-JEPA learning framework, which allows us to instantiate a Joint-Embedding Predictive Architecture in the context of images using masking.

\textbf{Masked Image Modeling:} an input image is first tokenized into patch embeddings $
\{e_i^p\}_{i=1}^N$, as explained in Section~\ref{scott}. Following that, a portion of the patch embeddings is selected to be masked. Denoting the masked position set as $M$, a shared learnable embedding $e_M$ replaces the original patch embeddings $e_i^p$ when $i\in M$, producing the masked sequence:
\begin{equation}
    e_i^M=\delta (i{\in}M) \odot  e_M+(1-\delta
(i{\in}M)) \odot e_i^p
\end{equation}
where  $\delta(\cdot)$ is the indicator function. Subsequently, the positional embedding is added and then fed the sequence into the $L$ transformer encoder blocks. After that, the output vectors $s=\{s_i\}_{i=1}^N$ are regarded as the encoded semantic representations of the input image patches. Thus, $s_i$ is the representation associated with the $i^{th}$ patch.

\textbf{Learning Image Representations in a Joint-Embedding Predictive Architecture through Masked Image Modeling (MIM-JEPA).} JEPAs are conceptually close to Generative Architectures, however, the loss function is applied in embedding space, not input space. The overall training objective is as follows: given a masked image as context to a context-encoder, task a predictor to learn the latent representations of the masked patches of the image produced by a target-encoder that is fed with the full image. We use a SCOTT enabled ViT, introduced in Section~\ref{scott}, for the context-encoder $f_\theta$  and target-encoder $f_{\bar{\theta}}$, the predictor $f_\phi$ is a shallow standard transformer \citep{vaswani2017attention} that takes as input the context-encoder outputs. Following, we describe how we produce each of the MIM-JEPA components: masking, targets, context, prediction, and loss, given an input image.

\textbf{Masking.}  In order to generate the masks for our MIM objective, we follow previous work to use a Blockwise masking strategy \citep{bao2021beit}. Specifically, given an input image, we iteratively sample possibly overlapping blocks with a random aspect ratio until enough patches are masked in $M$. In our experiments, $0.6N$, where $N$ is the total number of patches and 0.6 is the masking ratio. This masking strategy produces masked context-images that are informative and target-patches that are relatively semantic. See the masked image,  $I_{masked}$, in Figure \ref{fig:method}.

\textbf{Targets.} In the MIM-JEPA framework, the targets correspond to the latent representations of image blocks $s^y=\{s_i^y\}_{i=1}^N$ produced by the target-encoder $f_{\bar{\theta}}$ fed with the full input image, $I_{full}$. Thus, once $s^y$ is available, the target blocks are obtained by masking $s^y$ instead of the input image.

\textbf{Context.} Similarly, the masked input image $I_{masked}$, i.e., the image with patch-size holes, see Figure \ref{fig:method}, is fed into the context-encoder network $f_\theta$ to produce the corresponding patch-level representation $s^x=\{s_i^x\}_{i=1}^N$.

\textbf{Prediction.} Since the goal behind JEPAs is to predict the representations in an embedding space, we feed the context patch-level representations $s^x$ to the predictor $f_\phi$, which outputs the corresponding patch-level predictions $\hat  s^y=\{\hat 
s_i^y\}_{i=1}^N$.

\textbf{Loss.} The loss $L$ is simply the Smooth-L1 loss over the predictions $\hat s^y$ and the $N$ layer normalized \citep{ba_layer_2016} features $s^y$  produced by the target-encoder $f_{\bar \theta}$. Importantly, the loss is only applied to the masked patches to encourage the model to learn patch-level representations that are predictive of each other; predicting non-masked patches is trivial. 

The full training objective can be unified as:

\begin{equation}
    MIM=L(f_\phi(f_\theta(I_{masked})),N(f_{\bar{\theta }}(I_{full})))
\end{equation}

The parameters of the context-encoder, $\theta$, and the predictor, $\phi$, are jointly learned via gradient-based optimization, while the target-encoder’s parameters, $\bar\theta$, are updated via an exponential moving average (EMA) of the context-encoder parameters. Using an EMA target-encoder, an asymmetric architecture between the $x$- and $y$- encoding paths, and the layer normalization over target features $s^y$ has proven to avoid representation collapse and help training in previous works \citep{assran2023self, grill2020bootstrap, balestriero_cookbook_2023}, the same holds true for MIM-JEPA.

\textbf{Image augmentations.} Drawing inspiration from view-invariant SSL methods, we try to induce a shape-bias--a property of human perception \citep{naseer2021intriguing}--by randomly applying a set of simple image transformations: color jitter, grayscale, and gaussian blur, to a given input image to produce two views with slightly different color properties while preserving  spatial content.

\section{Experiments}
\label{sec:experiments}

\subsection{Datasets}
Recall that our objective is to develop a method capable of efficiently training from scratch on small-sized, standard-resolution, fine-grained datasets, while still maintaining the in-domain state-of-the-art representational power of bigger foundational models. In that sense, we conduct quantitative and qualitative studies on 3 datasets, 2 popular computer vision datasets from the VTAB benchmark \citep{zhai_large-scale_2020}: Oxford Flowers-102 \citep{nilsback_automated_2008} and Oxford IIIT Pets-37 \citep{parkhi_cats_2012}, and the ImageNet-100 \citep{deng_imagenet}. We selected these datasets for several reasons: (i) they are all considered small-sized datasets in the literature with a huge gap in Top-1 accuracy between from-scratch training and large-scale pretrained models \citep{steiner_how_2022}, (ii)  they are all standard-resolution, i.e., $224^2$ images. (iii) All three datasets present a significant challenge due to their high intra-class similarity. (iv) ImageNet-100 is a subset of ImageNet that contains 100 different classes of animals. (v) The magnitude of the image-per-class ratio for supervised training increases across the datasets, where $ratio=\frac{I_{train}}{N_{classes}}$. 

\begin{itemize}
    \item \textbf{Oxford Flowers-102} \citep{nilsback_automated_2008} The task consists in classifying among images of flowers present in the UK (102 classes, with between 40 and 248 images per class) with a total of 2,040 images for training (1,020 as validation split) and 6,149 for evaluation. Each image dimension has at least 500 pixels.
    \item \textbf{Oxford IIIT Pets-37} \citep{parkhi_cats_2012} The task consists in classifying images of dog and cat breeds (37 classes, with around 200 pictures each). The domain-specific features challenges models to differentiate between breeds that may be visually similar. There are 3,680 images for training and 3,669 for testing.
    \item \textbf{ImageNet-100} \citep{deng_imagenet} The task consists in classifying images of 100 different classes of animals present in the well-studied ImageNet dataset. There are 130,000 images for training (with roughly 1300 images per class) and 5,000 images for testing.
\end{itemize}

\subsection{Self-supervised Pretraining (SCOTT + MIM-JEPA)}

\label{subsec:ssl-scott-mim-jepa}
In contrast to supervised learning (SL), which requires labeled datasets, our MIM-JEPA pretraining strategy enables models to harness the full power of unsupervised learning paradigms by learning representations directly from the data itself, without labels (i.e., no label-based learning signal is involved during this pre-training stage). 

\begin{figure}[]
    \centering
    \includegraphics[width=\textwidth]{"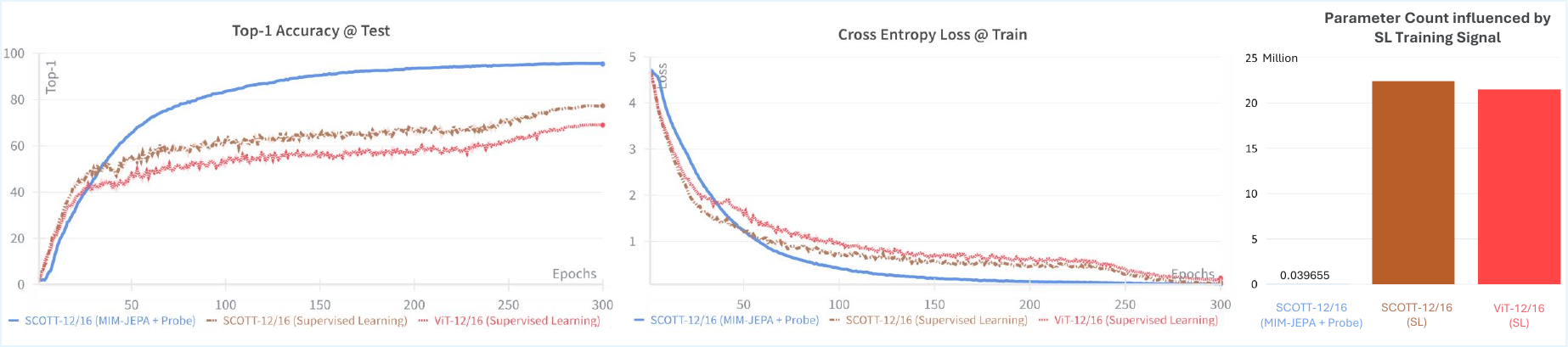"}
    \caption{\textbf{Training dynamics under supervised vs. self-supervised (MIM-JEPA) setups}. Evolution across 300 epochs on Flowers-102 of (left) Top-1 accuracy on the test set, (middle) cross-entropy loss on the training set, and (right) number of parameters directly influenced by the supervised learning signal. SCOTT-12/16 pretrained with MIM-JEPA and evaluated via a lightweight linear probe ($\approx$39k parameters) learns features that significantly outperform both fully supervised patch-and-embed tokenizer ViT-12/16 and dense convolutional tokenizer SCOTT-12/16 ($>$21M parameters). While supervised learning achieves higher accuracy early in training, self-supervised features generalize better and achieve substantially higher final test accuracy despite converging to similar training loss.}
    \label{fig:training}
\end{figure}

To further strengthen the comparison with supervised baselines \citep{xiao2021early, dosovitskiy_image_2021}, we analyze training dynamics over 300 epochs (Figure~\ref{fig:training}). Four main observations emerge: (i) supervised training initially achieves higher accuracy in the early epochs, but the self-supervised MIM-JEPA features steadily overtake and ultimately reach a substantially higher Top-1 accuracy on the test set; (ii) all methods converge to a similar cross-entropy loss on the training set, indicating that the observed generalization gap is not due to optimization difficulty; (iii) remarkably, a lightweight probe with only $\approx$39k trainable parameters on top of frozen MIM-JEPA features surpasses the performance of fully supervised models with over 21M parameters; and (iv) under identical MIM-JEPA training (same epochs, schedule, and evaluation), SCOTT consistently yields higher Top-1 accuracy than patchify, indicating improved sample efficiency at fixed compute. These findings are consistent with the broader evaluation across datasets reported in Tables~\ref{tab:SL_scratch}, \ref{tab:SL_finetuned} and \ref{tab:ablations}, confirming that SCOTT + MIM-JEPA not only reduces the reliance on large label-based signals but also yields more generalizable representations than conventional supervised training.

\textbf{Optimization}. All models are trained at $224\times 224$ input resolution. We use AdamW \citep{loshchilov2017decoupled} to jointly optimize the context-encoder and predictor with a batch size of 128, fitting in a single NVIDIA RTX 3090 GPU. For the learning rate, we follow a explore-exploit schedule \citep{iyer2023explore} with a linear warmup to its peak value of $5e-4$, a flat \textit{explore} phase for $0.72$ of the remaining epochs and a final \textit{exploit} phase with a cosine decay schedule. Weight decay is linearly increased from $0.04$ to $0.4$. For the target-encoder, the EMA parameter starts at $0.996$ and is linearly increased to $1$ during training. All hyperparameters are summarized in \ref{app:hyperparameters}.

\subsection{Downstream task: Frozen Image Classification}
\label{sec:frozen-classification}

To quantitatively demonstrate that our SCOTT + MIM-JEPA approach learns robust and highly semantic representations with scarce data and resources, we evaluate the transferability of frozen pretrained features on downstream classification tasks. Classification is selected due to its relevance in industrial and medical settings, where robust and interpretable representations are crucial for tasks such as defect or disease detection \citep{huang_self-supervised_2023}. Specifically, our experiments address three critical research questions:

\begin{enumerate}[label=(\Alph*)]
    \item How does a lightweight classifier trained on frozen SCOTT + MIM-JEPA features compare to fully supervised models trained from scratch?
    \item Can lightweight classifiers trained on frozen SCOTT + MIM-JEPA features achieve comparable accuracy to ViTs fine-tuned from large-scale supervised pretraining?
    \item How does our method compare to foundational state-of-the-art SSL methods pretrained on significantly larger and more diverse datasets?
\end{enumerate}

\textbf{Evaluation}. To answer these questions, after self-supervised pretraining (MIM-JEPA) on the unlabeled dataset for 300 epochs following Section \ref{subsec:ssl-scott-mim-jepa}, the model weights are frozen, and a simple, lightweight classifier ($\approx$35k params) is trained on top for $100$ epochs in a supervised manner. The images are resized to $256^2$ pixels from which a $224^2$ center crop is extracted. For all datasets, we report Top-1 and Top-5 classification accuracy as our main metrics. In addition, the experiments are repeated with 3 different random seeds, and we report results as mean~$\pm$~standard deviation. Random seeds are fixed across Python, NumPy, and PyTorch (CPU and GPU), deterministically controlling model initialization, mask generation, and data shuffling and augmentations to ensure reproducibility.

\textbf{A. Comparison with supervised training from scratch}. Table \ref{tab:SL_scratch} presents a performance comparison between SCOTT models pretrained using MIM-JEPA and Vision Transformers trained in a fully supervised manner from scratch for each target dataset. The results clearly demonstrate that our proposed approach significantly outperforms fully supervised training from scratch across all evaluated datasets and architectures. For instance, on the Pets-37 dataset, a ViT-12/16 trained from scratch achieves a Top-1 accuracy of 48.3\%, whereas an attentive probe on top of a frozen SCOTT-12/16 model pretrained with MIM-JEPA achieves a Top-1 accuracy of 90.7\% —an absolute improvement of 42.4 percentage points. Similarly, on the Flowers-102 dataset, supervised training from scratch yields a Top-1 accuracy of 79.1\% for SCOTT-12/16, while our method achieves a significantly higher 97.7\% with frozen pretrained features —an absolute improvement of 18.6 percentage points. This clearly illustrates the effectiveness and efficiency of SCOTT + MIM-JEPA features compared to traditional supervised training from scratch. Additionally, SCOTT-enabled ViTs outperform the standard ViT architecture in fully supervised learning. 

\begin{table}[]
\caption{Comparison of classification accuracy between Vision Transformers trained in a fully supervised manner from scratch and lightweight classifiers trained on frozen SCOTT features pretrained with MIM-JEPA. Evaluations are performed on fine-grained datasets (Flowers-102, Pets-37). Models marked with "*" are single runs of 1200 epochs.}
\label{tab:SL_scratch}
\centering
\resizebox{\textwidth}{!}{
\begin{tabular}{lccccc}
\hline
\textbf{}     & \multicolumn{1}{l}{} & \multicolumn{2}{c}{\textbf{Flowers-102}} & \multicolumn{2}{c}{\textbf{Pets-37}} \\ \cline{3-6} 
\textbf{Architecture} & \textbf{Params} & \textbf{Top-1}         & \textbf{Top-5}         & \textbf{Top-1}          & \textbf{Top-5}         \\ \hline
\rowcolor[HTML]{EFEFEF} \multicolumn{6}{l}{\textit{Supervised learning from scratch}}                                                                              \\
ViT-12/16     & 22 M                 & 71.1$\pm$0.04              & 87.3$\pm$0.12             & 48.3$\pm$0.02                    &  78.5$\pm$0.05                 \\
SCOTT-7/16    & 14 M                 & 79.0$\pm$0.09                   & \textbf{92.2$\pm$0.06}                   & 67.2$\pm$0.15                    & 89.3$\pm$0.09                   \\
SCOTT-12/16   & 22 M                 & \textbf{79.1$\pm$0.14}          & 91.9$\pm$0.04        & \textbf{67.4$\pm$0.09}          & \textbf{90.5$\pm$0.22}          \\ \hline
\rowcolor[HTML]{EFEFEF} \multicolumn{6}{l}{\textit{Frozen evaluation on top of MIM-JEPA pretraining.}}                                                      \\
SCOTT-7/16    & 14 M                 & 95.7$\pm$0.02                    & 99.1$\pm$0.07                   & 81.6$\pm$0.07                    & 97.4$\pm$0.15                    \\
SCOTT-12/16   & 22 M                 & 97.1$\pm$0.01                  & 99.1$\pm$0.01                  & 86.2$\pm$0.03                   & 98.5$\pm$0.02                    \\
SCOTT-7/16*   & 14 M                 & 96.9                   & \textbf{99.3}          & 88.0                    & 99.0                   \\
SCOTT-12/16*  & 22 M                 & \textbf{97.7}          & 99.2                   & \textbf{90.7}           & \textbf{99.4}          \\ \hline
\end{tabular}
}
\end{table}

\textbf{B. Comparison with fine-tuned ViTs from large-scale supervised pretraining}. Table \ref{tab:SL_finetuned} highlights that lightweight classifiers trained on top of frozen SCOTT + MIM-JEPA features achieve classification accuracy comparable to or superior to fully fine-tuned ViT models pretrained on large-scale labeled datasets. Notably, our method achieves these competitive resutls with significantly fewer parameters and drastically less data, demonstrating exceptional data-efficiency. For instance, on the Flowers-102 dataset, our SCOTT-7/16* model (14 M parameters) pretrained exclusively on just 8,189 unlabeled images achieves a Top-1 accuracy of 96.9\%, surpassing the 95.7\% accuracy of a larger ViT-12/16* model pretrained only on ImageNet-1K (1.3 million labeled images). Similarly, on Pets-37, our SCOTT-12/16* model pretrained on only 7,349 unlabeled images reaches a Top-1 accuracy of 90.7\%, closely approaching the performance of ViT models pretrained on vastly larger labeled datasets like ImageNet-1K (93.8\%) and ImageNet-21K (93.2\%). Moreover, on ImageNet-100, SCOTT-12/16 achieves a Top-1 accuracy of 84.8\%, competitive with SparseSwin pretrained on ImageNet-1K (86.9\%). These results underscore that SCOTT + MIM-JEPA achieves highly competitive performance while significantly reducing both computational resources and the amount of training data required.

\begin{table}[]
\caption{Comparison of classification accuracy achieved by lightweight classifiers trained on frozen SCOTT + MIM-JEPA features (pretrained exclusively on small target datasets) versus literature ViT models pretrained on large-scale labeled datasets (ImageNet-1K and ImageNet-21K) and fine-tuned end-to-end. Results demonstrate the data efficiency and competitiveness of our approach on fine-grained and medium-scale classification benchmarks. "\dag" denotes models pretrained exclusively on the unlabeled images from the respective target dataset. Models marked with "*" where pre-trained for 1200 epochs.}
\label{tab:SL_finetuned}
\centering
\resizebox{\textwidth}{!}{
\begin{tabular}{llccc}
\hline
\textbf{Architecture} & \textbf{Dataset (Size)} & \textbf{Flowers-102} & \textbf{Pets-37} & \textbf{ImageNet-100} \\ \hline
\rowcolor[HTML]{EFEFEF} \multicolumn{4}{l}{\textit{Fine-tuned ViTs from large-scale supervised pretraining.}}                            & \multicolumn{1}{l}{}  \\
ViT-12/16     & ImageNet-1K (1.3M)                  & 95.7                        & \textbf{93.8}                &              -         \\
ViT-12/16     & ImageNet-21K (14.2M)                & \textbf{99.6}               & 93.2                         &               -        \\
SparseSwin    & ImageNet-1K (1.3 M)                 & -                  & -                   & \textbf{86.9}         \\
DeiT-B \citep{touvron2021training}   & ImageNet-1K (1.3 M)                 & 98.5                  & -                   & -        \\
\hline
\rowcolor[HTML]{EFEFEF} \multicolumn{4}{l}{\textit{Frozen evaluation on top of MIM-JEPA pretraining.}}                                   & \multicolumn{1}{l}{}  \\
SCOTT-7/16   &               \multicolumn{1}{c}{\dag}                     & 95.7$\pm$0.02                       & 81.6$\pm$0.07                        & 81.1$\pm$0.11     \\
SCOTT-12/16   &              \multicolumn{1}{c}{\dag}                       & 97.1$\pm$0.01                        & 86.2$\pm$0.03                         & \textbf{84.8$\pm$0.12}        \\
SCOTT-7/16*   &               \multicolumn{1}{c}{\dag}                     & 96.9                        & 88.0                         & -       \\

SCOTT-12/16*  &                  \multicolumn{1}{c}{\dag}                   & \textbf{97.7}               & \textbf{90.7}                &            -           \\ \hline
\end{tabular}}
\end{table}

\textbf{C. Comparison to Foundational SSL Models}. Table \ref{tab:SSL} compares SCOTT models pretrained with MIM-JEPA against state-of-the-art foundational self-supervised learning methods such as DINOv2 \citep{oquab_dinov2_2024} and I-JEPA \citep{assran2023self}. These comparisons are reported primarily to contextualize scale; the core focus of this work is that our models are trained from scratch using only in-domain data, and achieve competitive performance under limited data and compute.

Remarkably, our approach achieves highly competitive performance despite using drastically smaller models and datasets. For instance, on the Pets-37 dataset, our SCOTT-12/16* model (22M parameters) pretrained exclusively on 7,349 unlabeled images achieves a Top-1 accuracy of 90.7\%, closely matching the performance (91.7\%) of I-JEPA's substantially larger ViT-32/14 model (630M parameters) pretrained on ImageNet-1K (1.3 million unlabeled images)—representing nearly 30 times fewer parameters and approximately 200 times less data. Similarly, on Flowers-102, our SCOTT-12/16* model achieves a superior Top-1 accuracy of 97.7\%, exceeding the performance (93.7\%) of I-JEPA's much larger ViT-32/14 model, again highlighting our method's exceptional data and parameter efficiency. Lastly, on ImageNet-100, our SCOTT-12/16* model pretrained on only 135,000 unlabeled images achieves a competitive Top-5 accuracy of 97.5\%, closely approaching the 98.9\% accuracy of DINOv2 pretrained on the vast LDV dataset (142 million unlabeled images).

\begin{table}[]
\caption{Contextual evaluation of frozen SCOTT + MIM-JEPA features against state-of-the-art foundational self-supervised learning methods (DINOv2, I-JEPA) pretrained on significantly larger and more diverse datasets. Our method achieves comparable accuracy despite using considerably fewer parameters and substantially less pretraining data and resources. "\dag" denotes models pretrained exclusively on the unlabeled images from the respective target dataset. SCOTT models were pretrained for 1200 epochs on Flowers-102 and Pets-37, while 300 epochs on ImageNet-100 due to dataset size.}
\label{tab:SSL}
\centering
\resizebox{\textwidth}{!}{
\begin{tabular}{lllcccccc}
\hline
\textbf{}           & \textbf{}         &                                        & \multicolumn{2}{c}{\textbf{Flowers-102}} & \multicolumn{2}{c}{\textbf{Pets-37}} & \multicolumn{2}{c}{\textbf{ImageNet-100}} \\
\textbf{Architecture}       & \textbf{Features} & \textbf{Dataset (Size)}                & \textbf{Top-1}      & \textbf{Top-5}     & \textbf{Top-1}    & \textbf{Top-5}   & \textbf{Top-1}      & \textbf{Top-5}      \\ \hline
\rowcolor[HTML]{EFEFEF} \multicolumn{9}{l}{\textit{Large-scale SSL pretrained methods.}}                                                                                                                                               \\
ViT-12/14+reg (22M) & DinoV2            & LVD (142M)                             & \textbf{99.6}                & \textbf{99.9}               & \textbf{94.8}              & \textbf{99.9}             & \textbf{89.1}                & \textbf{98.9}                \\
VIT-32/14 (630M)    & I-JEPA            & \multicolumn{1}{c}{ImageNet-1k (1.3M)} & 93.7                & 98.5               & 91.7              & 99.2             & 88.7                & 98.6                \\ \hline
\\ 
SCOTT-7/16         & MIM-JEPA          &                \multicolumn{1}{c}{\dag}                        & 96.9                & \textbf{99.3}           & 88.0              & 99.0             & 81.1                & 96.0                \\
SCOTT-12/16        & MIM-JEPA          &                   \multicolumn{1}{c}{\dag}                     & \textbf{97.7}                & 99.2             & \textbf{90.7}              & \textbf{99.4}             & \textbf{84.8}                & \textbf{97.5}                \\ \hline
\end{tabular}}
\end{table}

The competitiveness of SCOTT + MIM-JEPA, despite significantly fewer parameters and substantially less pre-training data, can be attributed to several key factors: (i) targeted pretraining directly on the domain-specific dataset, enabling the learning of highly specialized and semantically rich features; (ii) architectural optimization, where the SCOTT tokenizer may enable better representational efficiency and scalability compared to the traditional patch-and-embed strategy of ViTs; and (iii) the abstract latent self-supervised learning objective of MIM-JEPA, enabling the learning of robust and generalizable representations even from limited data. Furthermore, in contrast to most generative SSL frameworks that typically require fine-tuning all model parameters, our learning framework produces robust off-the-shelf features that enable the training of simple classifiers on top.

Importantly, ablations in Sec. \ref{sec:ablations} show that these gains are not explained solely by enforcing mask/occlusion invariance (MAE / C-MAE \citep{he2022masked,kong2023understanding}), further supporting the role of latent predictive learning (MIM-JEPA) and sparse convolutional tokenization (SCOTT) in low-data regimes.

This property of discriminative SSL \citep{caron2021emerging, oquab_dinov2_2024} is achieved in our setup without complex image augmentations to introduce view-invariant biases. These attributes make SCOTT + MIM-JEPA particularly attractive for practical deployment in scenarios where labeled data, computational resources, and extensive training infrastructure are limited, offering promising pathways for future exploration into broader application domains and more diverse datasets.

\subsection{Efficiency and Resource Profile}

To complement our accuracy results, we report an efficiency comparison between SCOTT and a ViT baseline with an identical transformer backbone. Because SCOTT is a drop-in replacement for the standard patch-embed layer, efficiency arise solely from this component. Unlike the dense patch-embed in ViT, which computes features for all patches regardless of masking, SCOTT levarages sparse convolutions whose FLOPs scale linearly with the visible patch fraction, making it more efficient at higher mask ratios. Table \ref{tab:flops} reports tokenizer FLOPs and MACs, as well as measured end-to-end runtime metrics (single-image latency, batch-32 throughput, and peak GPU memory) on an NVIDIA RTX 3090.

Although SCOTT introduces a higher tokenizer compute cost (2.13G FLOPs vs.\ 115.7M FLOPs for ViT patch-embed), the observed runtime overhead is modest: inference latency increases only from 8.96 to 9.86~ms per image (batch size~1), and throughput remains high (1158~img/s vs.\ 1360~img/s at batch size~32). Peak GPU memory usage shows only a small increase (0.18 vs.\ 0.17~GB at b=1), confirming that SCOTT achieves its superior representation quality without prohibitive runtime or memory penalties. Full FLOPs and extended efficiency analyses are provided in \ref{app:efficiency}.

\section{Building intuitions with ablations}
\label{sec:ablations}

We conduct ablation studies to isolate the individual contributions of the SCOTT tokenizer and the MIM-JEPA pretraining objective, and to contextualize their behavior against representative masked pretraining baselines. Unless stated otherwise, we use a SCOTT-12/16 backbone (comparable in parameter count to ViT-S) and follow the same frozen probing protocol used throughout the paper. All models are pretrained for 300 epochs, consistent with the shorter pretraining runs used in our main results, and evaluated on two standard fine-grained benchmarks: Flowers-102 and Pets-37. These datasets provide complementary low-data regimes: Flowers-102 contains few labeled samples per class and high intra-class similarity, while Pets-37 includes visually similar sub-categories and a larger training set. Results are summarized in Table~\ref{tab:ablations}; While we focus Table \ref{tab:ablations} on Flowers/Pets due to their extreme low-data character, ImageNet-100 shows consistent trends (Table \ref{tab:SL_finetuned}). additional ablations are provided in Appendix~\ref{app:ablations}.

\begin{table}[!hbt]
\caption{Ablation studies for SCOTT models and MIM-JEPA pretraining on Flowers-102 and Pets-37. MAE~\citep{he2022masked} and its occlusion-invariant variant C-MAE~\citep{kong2023understanding} are included as strong masked pretraining baselines with different objectives and are evaluated under the same pretraining budget and frozen probing protocol.}

\label{tab:ablations}
\centering
\resizebox{\textwidth}{!}{
\begin{tabular}{lcccc}
\hline
\multirow{2}{*}{\textbf{Models}}                      & \multicolumn{2}{c}{\textbf{Flowers-102}} & \multicolumn{2}{c}{\textbf{Pets-37}}                        \\
                                                      & \textbf{Top-1}            & \multicolumn{1}{c}{\textbf{Top-5}} & \textbf{Top-1}            & \multicolumn{1}{c}{\textbf{Top-5}} \\ \hline
SCOTT-12/16 and MIM-JEPA pretraining (300 Epochs). \textit{(ours)}     & \textbf{97.15}    & \multicolumn{1}{c}{\textbf{99.15}}    & \textbf{90.7}    & \multicolumn{1}{c}{\textbf{99.4}}    \\ \hline
- No MIM-JEPA \& No SCOTT (i.e., ViT-12/16 supervised learning) & 71.08                     & \multicolumn{1}{c}{87.52}   & 48.3  & \multicolumn{1}{c}{78.5}        \\
- SCOTT Tokenizer without MIM-JEPA pretraining. & 79.13                     & \multicolumn{1}{c}{91.96}   & 67.4                  & \multicolumn{1}{c}{90.5} \\
- MIM-JEPA pretraining without SCOTT Tokenizer (ViT-12/16)     & 95.25 & 99.07 & 73.61                   & \multicolumn{1}{c}{94.76}                     \\

- MAE \citep{he2022masked} pretraining (ViT-12/16) & 
81.16 & 94.86 & 64.18                   & \multicolumn{1}{c}{91.41}    \\

- C-MAE \citep{kong2023understanding} occlusion-invariant pretraining (ViT-12/16) & 76.41 & 90.76  & 44.23  & \multicolumn{1}{c}{78.63}     \\ \hline
\end{tabular}}
\end{table}

\textbf{Mask-invariance baselines (MAE / C-MAE).}
We report MAE~\citep{he2022masked} and C-MAE~\citep{kong2023understanding} as two strong masked pretraining baselines with different learning objectives. MAE is a reconstruction-based MIM method that predicts masked patches in pixel/patch space using a decoder, whereas C-MAE replaces pixel reconstruction with a siamese representation learning objective that enforces occlusion invariance by contrasting complementary masks in feature space (InfoNCE). Both baselines are evaluated under the same pretraining budget and frozen probing protocol, focusing on the quality of off-the-shelf representations.

On Flowers-102, MAE reaches 81.16\% Top-1, while C-MAE decreases to 76.41\%. On Pets-37, MAE improves to 64.18\%, whereas C-MAE collapses to 44.23\% Top-1, comparable to supervised learning from scratch. Under this frozen evaluation, neither reconstruction-based training (MAE) nor explicit occlusion-invariant feature learning (C-MAE) yields linearly separable representations competitive with latent predictive learning. In contrast, MIM-JEPA produces substantially stronger features (95.25\% on Flowers-102 and 73.61\% on Pets-37), and SCOTT + MIM-JEPA further improves performance on both datasets (97.15\% and 90.7\% Top-1).

Notably, Top-5 accuracy is generally closer across methods, whereas Top-1 accuracy reveals large differences in class separability, reinforcing that the improvements stem from representation quality rather than coarse semantic grouping.

\textbf{Self-supervised pretraining is critical in low-data regimes.}
Table~\ref{tab:ablations} highlights the large gap between training from scratch with labels and learning representations through self-supervised pretraining. In fully supervised training, ViT-12/16 achieves only 71.08\% Top-1 on Flowers-102 and 48.3\% on Pets-37, while SCOTT-12/16 provides modest gains (79.13\% and 67.4\%), suggesting that architectural priors alone are insufficient when the model must learn both low-level structure and semantics from limited labeled data. In contrast, frozen features pretrained with MIM-JEPA enable substantially stronger performance under a lightweight classifier, reaching 97.15\% on Flowers-102 and 90.7\% on Pets-37. This gap demonstrates that the dominant factor in achieving high performance under limited labels is the ability to learn transferable representations from unlabeled data.

\textbf{Why does MIM-JEPA improve few-sample training?}
A key challenge in low-data self-supervised learning is not only to learn meaningful representations, but to learn representations that are immediately usable without end-to-end adaptation. Reconstruction-based MIM objectives optimize for recovering missing content in pixel or patch space, which can emphasize low-level variability and does not necessarily yield features that separate fine-grained categories under a lightweight probe. Similarly, contrastive learning with complementary masks explicitly promotes occlusion invariance, but may still underutilize latent predictive structure when trained from scratch on limited data.
MIM-JEPA instead performs masked prediction directly in representation space by aligning the context-encoder outputs with targets produced by a momentum target encoder. This latent-space objective promotes semantic structure while reducing the need to reconstruct pixel-level details, leading to representations that are more discriminative under frozen evaluation.
This behavior is consistent with the observed gap between MAE and MIM-JEPA on both datasets (Flowers-102: 81.16$\rightarrow$95.25; Pets-37: 64.18$\rightarrow$73.61 Top-1).

\textbf{Why does SCOTT improve few-sample training?}
We next isolate the effect of SCOTT by comparing MIM-JEPA pretraining with a standard ViT patch-and-embed stem. Under identical MIM-JEPA training, replacing patchify with SCOTT improves Top-1 accuracy on both Flowers-102 (95.25$\rightarrow$97.15) and Pets-37 (73.61$\rightarrow$90.7).
This suggests that SCOTT contributes more than a stronger patch embedding layer: it injects locality priors at the tokenization stage while remaining compatible with masked modeling.
In small-data regimes, patchify discretizes the image into independent tokens and removes continuity cues across patch boundaries; with limited pretraining data, attention layers may not reliably recover such priors. SCOTT mitigates this by using submanifold sparse convolutions that operate only on visible spatial sites, preventing feature contamination across masked boundaries and avoiding mask-vanishing effects that arise in dense convolutional stems. This provides a stable, masking-consistent inductive bias during tokenization, improving the linear separability of the learned features, particularly on the more challenging Pets-37 dataset.

\textbf{Effect of color augmentations.}
As shown in Appendix~\ref{tab:ablation-augmentations}, disabling color augmentations results in less than a 2-point drop on Flowers-102 (97.15\% $\rightarrow$ 95.86\%). This suggests that SCOTT + MIM-JEPA can learn robust representations without strong augmentation, which is important for domains where image-specific augmentations are limited or inappropriate (e.g., medical imaging or other sensing modalities).

\section{Qualitative Results}

\begin{figure}[!htb]
    \centering
    \includegraphics[width=\textwidth]{"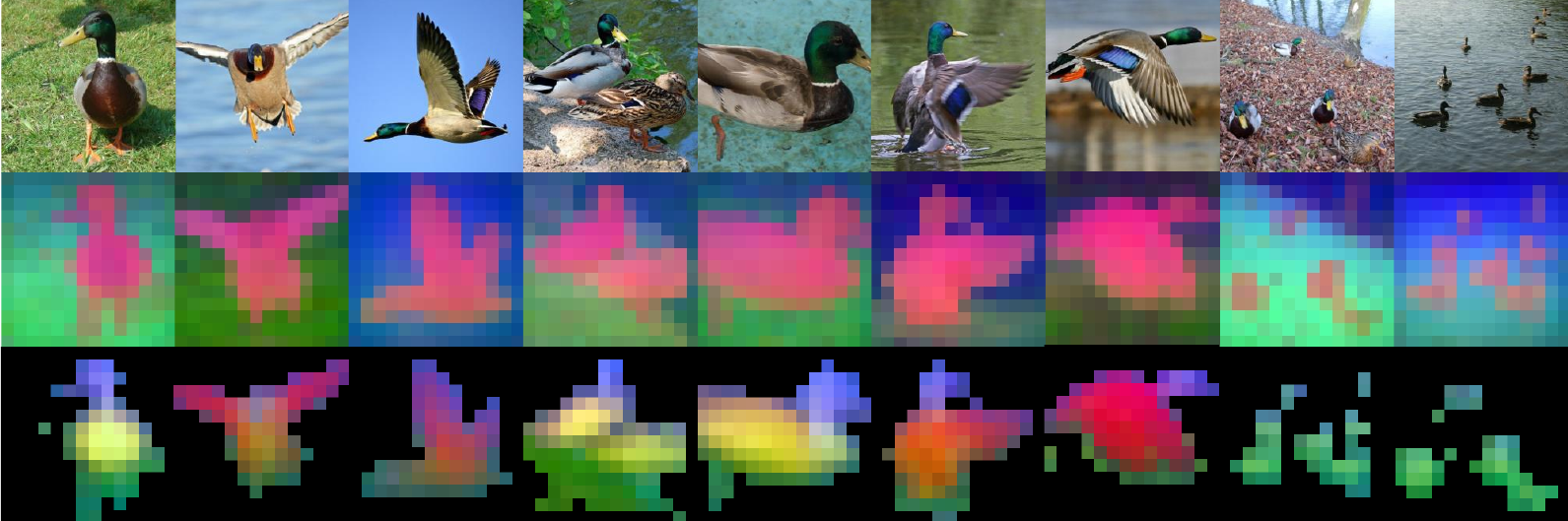"}
    \caption{\textbf{Visualization of semantic structures via PCA on SCOTT + MIM-JEPA frozen patch embeddings.}. The top row shows original images, while subsequent rows display PCA-based segmentation. The first PCA differentiates foreground (pink) from background through thresholding. A second PCA further distinguishes semantic object-parts consistently across images (e.g., heads in purple, wings in red, torsos in yellow). Similar to Figure \ref{fig:pca_many} (c), the columns on the right demonstrate the model's ability to identify and segment multiple instances, indicating potential utility in object detection and counting tasks.}
    \label{fig:pca_duck}
\end{figure}

To gain deeper insight into the semantic representations learned by our SCOTT + MIM-JEPA model, we perform qualitative visualizations through principal component analysis (PCA) on the learned frozen patch features. Results of this visualization are presented in Figure \ref{fig:pca_many} and Figure \ref{fig:pca_duck}. 

\textbf{PCA Visualization of Patch Features.} In Figure \ref{fig:pca_duck}, we visualize the first three principal components of patch embeddings produced by our model, mapping each component to the RGB color channels for intuitive interpretation. Remarkably, despite not being explicitly trained for semantic segmentation or object-part recognition, our model spontaneously generates semantically meaningful and consistent segmentations. Specifically, the first principal component effectively differentiates foreground (main object) patches from the background, enabling straightforward object segmentation through simple thresholding. Subsequent PCA on these foreground patches further reveals distinct semantic sub-components, such as heads, wings, and torsos, consistently across diverse examples within the same semantic \textit{family}. For example, in Figure \ref{fig:pca_duck}, ducks are segmented clearly from the background (pink regions), and their semantic parts—head (purple), wings (red), and torso (yellow)—are distinctly identified. Additionally, the capability of our model to segment multiple objects within a single image (evidenced by the two rightmost columns in Figure \ref{fig:pca_duck}) suggests potential for extending these learned representations to object detection and counting tasks.

This emerging property shared with other SSL methods was previously reported in DinoV2 \citep{oquab_dinov2_2024}. However, our method achieves comparable visual semantic parsing without relying on extensive image augmentations, large-scale pre-training, or specialized tokens (e.g., class tokens), underscoring the efficiency and robustness of the learned representations. Such intuitive and interpretable features make our approach particularly valuable for applications requiring explainability and interpretability, including medical diagnostics and automated quality inspections, where the interpretability of model decisions can be as crucial as raw performance metrics.

\section{Conclusion}

Effective representation learning in computer vision traditionally depends heavily on large-scale datasets and vast computational resources. In this work, we challenged this paradigm by demonstrating that powerful, off-the-shelf representations can be effectively learned with significantly fewer data samples, reduced computational demands, and smaller model sizes. By integrating a Sparse Convolutional Tokenizer into Transformer architectures, we inject CNN-like inductive biases while maintaining compatibility with masked image modeling through MIM-JEPA self-supervised pretraining. Our results clearly demonstrate that lightweight classifiers trained on frozen SCOTT + MIM-JEPA pretrained features significantly outperform fully supervised models trained from scratch, achieving performance competitive with state-of-the-art methods despite pretraining exclusively on small-scale datasets without complex augmentations. 

Beyond overall performance, our ablation studies provide additional insights into what enables data-efficient self-supervised learning in this setting. In particular, we find that standard masked autoencoding baselines and occlusion-invariant objectives (MAE and C-MAE) underperform MIM-JEPA, and remain far below SCOTT + MIM-JEPA, despite being evaluated with an identical probing protocol. This suggests that enforcing mask/occlusion invariance alone is not sufficient to escape the big-data paradigm in standard fine-grained regimes, and highlights the importance of combining latent-space prediction with sparsity-aware convolutional inductive priors that remain compatible with masking.

These findings highlight the potential of our method to a long tail of computer vision applications beyond natural images, particularly in domains with limited labeled data and constrained computational resources common scenarios in industrial automation, embedded systems, and medical imaging. Future research will investigate fine-tuning strategies, extension to dense prediction tasks such as semantic segmentation, and adaptation to highly specialized datasets. Continued exploration into methods like SCOTT + MIM-JEPA represents a promising direction toward democratizing deep learning, making robust visual recognition accessible beyond the conventional big data paradigm.

\subsubsection*{Acknowledgments}
\textbf{Funding.} This research was financed by the European Union through the European Regional Development Fund, within the Operational Programme of the Valencian Community 2021-2027 within the QRAIS project with grant number IMDEEA/2024/46.

 \bibliographystyle{elsarticle-num} 
 \bibliography{cas-refs}

\appendix
\section{Architecture}

\subsection{Sparse Convolutional Tokenizer for Transformers (SCOTT) Architecture}
\label{app:scott-arch}

\begin{table}[!htb]
\caption{Architecture of the Sparse Convolutional Tokenizer for Transformers (SCOTT)}
\centering
\resizebox{\textwidth}{!}{
\begin{tabular}{ccccccc}
\hline
\textbf{Layer} & \textbf{Type}         & \textbf{\# in} & \textbf{\# out} & \textbf{Kernel size} & \textbf{Stride} & \textbf{Padding} \\ \hline
1              & Sparse Convolution 2D & 3                    & 64                    & 7                    & 2               & 3    \\
2              & ReLU                  & -                    & -                     & -                    & -               & -    \\
3              & Sparse MaxBlurPool 2D & -                    & -                     & 3                    & 2               & -    \\
4              & Sparse Convolution 2D & 64                   & 384                   & 7                    & 2               & 3    \\
5              & ReLU                  & -                    & -                     & -                    & -               & -    \\
6              & Sparse MaxBlurPool 2D & -                    & -                     & 3                    & 2               & -    \\ \hline
\end{tabular}}
\end{table}

\subsection{Transformer backbone}

\begin{table}[!hbt]
\caption{SCOTT Transformer backbone variants}
\centering
\resizebox{\textwidth}{!}{
\begin{tabular}{lcccccc}
\hline
\textbf{Model} & \textbf{Emb. Dim.} & \textbf{Pos. Emb.} & \textbf{\# Blocks} & \textbf{\# Heads} & \textbf{FFN} & \multicolumn{1}{l}{\textbf{\# Params}} \\ \hline
SCOTT-7/16     & 384                & Fixed   & 7                 & 4                & SwiGLU       & 13.6 M                                \\
SCOTT-12/16    & 384                & Fixed   & 12                & 6                & SwiGLU       & 22.4 M                                \\ \hline
\end{tabular}}
\end{table}

\subsection{FLOPs Computation and Extended Efficiency Analysis}
\label{app:efficiency}
Because the only architectural difference between SCOTT and a standard ViT is the replacement of the patch-embed stem with the SCOTT tokenizer, efficiency differences arise solely from this component. Table~\ref{tab:flops} reports both compute for the tokenizer stage (FLOPs and MACs) and measured end-to-end inference metrics.

\begin{table}[!hbt]
\caption{Tokenizer compute and end-to-end inference resource profile comparing SCOTT and ViT equivalent on 224$\times$224 input (NVIDIA RTX 3090). FLOPs/MACs refer to the \emph{tokenizer only}; latency/throughput and peak memory are measured end-to-end. We count 1 MAC = 2 FLOPs.}
\label{tab:flops}
\centering
\resizebox{\textwidth}{!}{
\begin{tabular}{lcccccc}
\hline
\textbf{Tokenizer} & \textbf{FLOPs} & \textbf{MACs} & \textbf{Latency} & \textbf{Throughput} & \textbf{Peak Mem. } & \textbf{Peak Mem.}\\ 
 & & & (b=1) & (b=32) & (b=1) & (b=32) \\  \hline
SCOTT-12/16               & 2.1314 G    & 1.0646 G  & 9.86 ms/img & 1158.52 img/s & 0.18 GB & 0.61 GB\\
ViT-12/16 (Patch-Embed)   & 115.6810 M  & 57.8028 M & 8.96 ms/img & 1360.27 img/s & 0.17 GB & 0.36 GB\\ 
\hline
\end{tabular}}
\end{table}

We compute FLOPs as the total number of floating-point multiply–accumulate operations (MACs) performed in a single forward pass, counting one MAC as two FLOPs (multiply and add). In ViT, the patch-embed stem applies a dense $16 \times 16$ convolution to all image patches, resulting in a constant cost of $\approx 115.6$M FLOPs per $224 \times 224$ image, regardless of masking. By contrast, SCOTT replaces this step with a shallow stack of sparse $7 \times 7$ convolutions. Because these convolutions operate only on active (visible) spatial sites, their FLOPs scale approximately linearly with the visible fraction $(1 - m)$, where $m$ is the mask ratio. This makes SCOTT naturally more efficient at higher mask ratios during training.

The right half of Table~\ref{tab:flops} presents the end-to-end runtime measurements on an NVIDIA RTX 3090: single-image latency, batch-32 throughput, and peak memory usage for both batch sizes. Despite SCOTT’s higher tokenizer-level FLOPs, the difference in end-to-end latency is modest ($\sim 1$~ms per image), and SCOTT maintains competitive throughput and memory footprint. These results demonstrate that SCOTT’s sparsity-aware design does not impose prohibitive runtime overheads while enabling stronger representations and scaling benefits.

\section{Image augmentations}

During self-supervised training, MIM-JEPA uses the following image augmentations to generate different views while preserving content location:
\begin{itemize}
    \item Random cropping: a random patch from the original image is selected with an area uniformly sampled between 0.2 and 1.0, and an aspect ratio between 3/4 and 4/3. Once cropped, the patch is resized using bicubic interpolation to the target size 224×224.
    \item 50\% chance of horizontal flip.
    \item Color jittering: random uniformly change the brightness (0.4), contrast (0.4), saturation (0.2), hue (0.1), with a probability of 0.8.
    \item Grayscale conversion with a probability of 0.1.
    \item Gaussian blurring: with a probability of 0.3 for a 224x224 image, apply a square Gaussian kernel of 9x9 and a standard deviation uniformly sampled between 0.1 and 2.
\end{itemize}
	
In the default pretraining strategy, each image view is generated through a different augmentation pipeline. First random cropping and horizontal flipping take place, then the order in which color jitter, grayscale and gaussian blurring augmentations are applied is uniformly sampled before applying the pipeline. Once that augmentation pipeline is applied, color channels are normalized by subtracting the average color and dividing by the standard deviation, computed on ImageNet.

\section{MIM-JEPA pretraining hyperparameters}
\label{app:hyperparameters}

\begin{table}[!hbt]
\centering
\caption{MIM-JEPA pretraining hyperparameters}
\label{tab:mim-jepa-hyperparams}
\begin{tabular}{l|c}
\hline
\textbf{Parameter}           & \textbf{Value}            \\ \hline
Predictor \# Blocks           & 3                         \\ \hline
Masking                      & Blockwise                 \\
Mask ratio                   & 0.6                       \\ \hline
Batch size                   & 128                       \\
Optimizer                    & \multicolumn{1}{l}{AdamW} \\
\# Epochs                     & 300                       \\
Learning rate start          & 0.000001                  \\
Learning rate peak           & 0.0005                    \\
Learning rate final          & 0.00001                   \\
Learning rate flat (\%)       & 72                        \\
\# Linear warmup epochs       & 40                        \\
Learning rate decay Schedule & Cosine                    \\
Weight decay start           & 0.04                      \\
Weight decay end             & 0.4                       \\
Weight decay Schedule        & Linear                    \\
EMA start                    & 0.996                     \\
EMA end                      & 1.0                       \\
EMA Schedule                 & Linear                   \\ \hline
\end{tabular}
\end{table}

SCOTT models that are pretrained for longer, i.e., 1200 epochs, also warmup for longer, i.e., 60 epochs. The rest of hyperparameters is kept the same as in Table \ref{tab:mim-jepa-hyperparams}.

\section{Evaluation}

\subsection{Evaluation protocols}
\label{app:eval-protocols}

Given an input image, the SCOTT model pretrained using MIM-JEPA outputs a sequence of features $s=\{s_i \}_{i=1}^N$, where  $s_i$ is the encoded semantic representation associated with the $i^{th}$ image patch. A feature pooling operation is applied to $s$ to generate a single feature vector, which is then fed into a linear classifier for downstream supervised tasks. Following literature, we report results obtained with two different pooling strategies: a linear operation (average pooling) and a non-linear operation (attentive pooling).

\textbf{Linear Probing}. To pool the sequence of features $s=\{s_i\}_{i=1}^N$ into a single vector, a simple linear operation (average pooling) is applied, followed by a LayerNorm. The resulting feature vector is fed into a linear classifier.

\textbf{Attentive Probing} \citep{bardes2024revisiting}. To pool the sequence of features $s=\{s_i \}_{i=1}^N$ into a single vector, a lightweigth non-linear cross-attention block with a learnable query token is learnt.  The output of the cross-attention block is added back to the query token through a residual connection and fed into a SwiGLU layer, followed by a LayerNorm. The resulting feature vector is fed into a linear classifier.

\subsection{Evaluation details}

Details regarding numbers reported in Section \ref{sec:frozen-classification}. For fair comparisons, unless stated otherwise, all methods share the same image augmentations and hyperparameters as presented in Table \ref{tab:mim-jepa-hyperparams}:

\begin{itemize}
    \item In Table \ref{tab:SL_scratch}, supervised ViTs and SCOTT variants are trained for 300 epochs.
    \item In Table \ref{tab:SL_finetuned}, fine-tuned ViTs are extracted from \citep{steiner_how_2022}. SparseSwim result is from \citep{pinasthika_sparseswin_2024}.
    \item In Table \ref{tab:SSL}, all self-supervised methods reported, i.e., DinoV2, I-JEPA, MIM-JEPA, are probed on best result after 100 epochs on the target dataset. DinoV2 uses a linear-probe on CLS token. Pretrained weights are publicly available. The ViT-12/14 is distilled from a ViT-g/14 (1,100 M parameters). I-JEPA uses an attentive-probe on patch tokens. Only pretrained weights for big model sizes (ViT-32/14) are publicly available.
    \item In Table \ref{tab:ablations}, MAE \citep{he2022masked} and occlusion-invariant MAE (C-MAE) \citep{kong2023understanding} baselines are evaluated using the same attentive probe on patch tokens, training schedule, compute budget, and data preprocessing as SCOTT + MIM-JEPA to ensure a fair comparison.
\end{itemize}

At the start of each run, we fix random seeds for Python, NumPy, and PyTorch (CPU and GPU), and enable deterministic cuDNN while disabling benchmarking. This ensures that seeds control model initialization, data shuffling, and mask generation in MIM. For each experiment, we repeat training with three different seeds and report mean ± standard deviation.

\section{Mask-invariance baselines (MAE/ C-MAE)}
\label{app:maskinv_baselines}

\textbf{MAE baseline.}
Masked Autoencoders (MAE) \citep{he2022masked} are a reconstruction-based masked image modeling framework in which an image is partitioned into non-overlapping patches and a large fraction of them are masked, with the model trained to reconstruct the missing content from the visible context. MAE consists of a ViT encoder operating on the visible (and masked) patch sequence and a lightweight transformer decoder that predicts the pixel-level content of the masked patches, using a mean squared error loss in normalized pixel space. In our implementation, we follow a MAE variant from \citep{kong2023understanding} in which learnable mask tokens are explicitly embedded in the encoder input (rather than dropping masked patches entirely), making the encoder pathway closer to our MIM-JEPA design while retaining the canonical MAE structure of a reconstruction decoder and pixel-space reconstruction loss.

\textbf{C-MAE baseline.} Contrastive-MAE (C-MAE) \citep{kong2023understanding} is an occlusion-invariant masked modeling method that replaces MAE’s decoder and pixel-space reconstruction loss with a siamese contrastive learning objective. Instead of predicting raw pixels, C-MAE forms two complementary masked views of the same image and processes them through an online encoder and a target encoder (EMA-updated), encouraging the resulting patch-level features to be consistent across masks. The learning signal is provided by a token-wise contrastive InfoNCE loss in feature space, applied between the representations of two complementary masks, with projection and prediction MLP heads to stabilize training. By optimizing agreement between complementary observations, C-MAE explicitly promotes occlusion-invariant features without requiring pixel reconstruction, making it a strong representative of mask-invariance objectives under masked pretraining.

\section{Ablations}
\label{app:ablations}

We conduct complementary ablations to those presented in Section \ref{sec:ablations} on Flowers-102.

\textbf{Masking strategy}. In Table \ref{tab:ablation-masking} we compare different masking strategies. Blockwise masking is our default strategy introduced in Section \ref{subsec:mim-jepa}. In random masking the target is a set of patches uniformly sampled from the encoded image representation . For both masking strategies, the context image is the complement of the masked target set, ensuring that there are no overlapping patches between the context and target blocks. Consistent with prior works, we find that MIM-JEPA benefits more from blockwise masking than from random masking. The intuition is that blockwise masking strikes a good balance in generating target blocks with relative semantic meaning while producing context blocks that are informative of the missing information. Additionally, higher masking ratios also improve performance.

\begin{table}[!htb]
\centering
\caption{Ablating masking strategy. Attentive and linear evaluation on Flowers-102 Dataset using the train split (2040 labeled samples) after MIM-JEPA pretraining of a SCOTT-12/16 enabled ViT for 300 epochs. Blockwise masking achieves superior performance in both attentive and linear evaluation. In addition, a higher masking ratio leads to better performance overall. }
\label{tab:ablation-masking}
\resizebox{\textwidth}{!}{
\begin{tabular}{lccccc}
\hline
\textbf{M strategy} & \textbf{M ratio} & \textbf{Top-1 Att.} & \textbf{Top-1 Lin.}   & \textbf{Top-5 Att.} & \textbf{Top-5 Lin.}    \\ \hline
Random     & 0.4     & 90.64           & 81.57          & 97.64           & 95.00          \\ 
Random     & 0.6     & 92.04           & 84.46          & 97.99           & 95.91          \\ 
Blockwise  & 0.4     & 95.85           & 92.66          & 98.86           & 98.38          \\ 
Blockwise  & 0.6     & \textbf{97.15}  & \textbf{94.81} & \textbf{99.15}  & \textbf{98.78} \\ \hline
\end{tabular}}
\end{table}

\textbf{Image augmentation strategy.} In the default MIM-JEPA pretraining strategy, we generate two (different) views of a given crop with a certain probability by slightly modifying only the color properties; thereby, preserving equivalent spatial content. We ablate the performance of this strategy versus applying the same color augmentation to both views (same) and to disabling color augmentations entirely (none). As shown in Table \ref{tab:ablation-augmentations}, (different) view augmentation strategy achieves best performance. However, it is noteworthy that the performance gap compared to using no augmentations (none) is less than 2 percentage points. This suggests that augmentations may not be necessary when pretraining SCOTT models within a MIM-JEPA framework. The intuition is that the JEPA objective of predicting in abstract representation space potentially mitigates the reliance on unnecessary pixel-level details. This is particularly relevant to fields (e.g., x-ray imaging) and modalities (e.g., audio) where image-specific augmentations are not feasible. 

\begin{table}[!htb]
\centering
\caption{Performance Comparison of Image Augmentation Strategies. The "different" view augmentation strategy achieves the highest performance across metrics. However, the performance gap compared to using no augmentations ("none") is less than 2 percentage points, suggesting that augmentations may not be necessary when pretraining SCOTT models within a MIM-JEPA framework.}
\label{tab:ablation-augmentations}
\resizebox{\textwidth}{!}{
\begin{tabular}{lcccc}
\hline
\textbf{Augmentation Strategy} & \textbf{Top-1 Attentive} & \textbf{Top-1 Linear}   & \textbf{Top-5 Attentive} & \textbf{Top-5 Linear}    \\ \hline
none                 & 95.86           & 92.60          & 98.82           & 97.82          \\ 
same                 & 96.76           & 94.29          & 99.12           & 98.56          \\ 
different            & \textbf{97.15}  & \textbf{94.81} & \textbf{99.15}  & \textbf{98.78} \\ \hline
\end{tabular}}
\end{table}

\begin{table}[!htb]
\caption{Performance comparison of SCOTT models with and without MIM-JEPA pretraining. The results demonstrate that MIM-JEPA pretraining significantly improves Top-1 accuracy by 18 percentage points compared to supervised training from scratch, even when only a lightweight classifier is trained on top of frozen pretrained weights.}
\label{tab:scott-no-mim-jepa}
\centering
\resizebox{\textwidth}{!}{
\begin{tabular}{llcccc}
\hline
\textbf{Strategy} & \textbf{Pretraining data (size)}  & \textbf{Top-1   Att.} & \textbf{Top-1   Lin.} & \textbf{Top-5   Att.} & \textbf{Top-5   Lin.} \\ \hline
None                   & -                              & 79.13             & 78.54          & 91.96             & 91.85          \\ 
MIM-JEPA               & Train   split (2040)           & 80.69             & 66.92          & 93.83             & 87.03          \\ 
MIM-JEPA   (ours)      & Train   + Test (8189)          & \textbf{97.15}             & \textbf{94.81 }         & \textbf{99.15}             & \textbf{98.78}          \\ \hline
\end{tabular}}
\end{table}

\begin{table}[!htb]
\caption{Performance comparison of MIM-JEPA pretraining with and without SCOTT Tokenizer. This table illustrates the importance of the SCOTT Tokenizer by comparing models where MIM-JEPA pretraining uses the standard patch embedding in ViT instead of the SCOTT Tokenizer. Notably, SCOTT-7/16 (13.6 M parameters) slightly outperforms ViT-12/16 (21.5 M parameters) despite having nearly half the parameters.}
\centering
\label{tab:mim-jepa-no-scott}
\resizebox{\textwidth}{!}{
\begin{tabular}{llcccc}
\hline
\textbf{Model}              & \textbf{\# Params}    & \textbf{Top-1 Att.} & \textbf{Top-1 Lin.}   & \textbf{Top-5 Att.} & \textbf{Top-5 Lin.}   \\ \hline
ViT-7/16           & 12.7 M       & 93.54           & 89.81          & 98.69           & 97.91          \\ 
ViT-12/16          & 21.5 M       & 95.25           & 92.82          & 98.78           & 98.40          \\ 
SCOTT-7/16         & 13.6 M       & 95.64           & 92.70          & 99.07           & 98.19          \\ 
SCOTT-12/16        & 22.4 M       & \textbf{97.15}  & \textbf{94.81} & \textbf{99.15}  & \textbf{98.78} \\ \hline
\end{tabular}}
\end{table}

\section{Scalability assessment}
\label{app:scalability}

High scalability is one of the primary advantages of the standard ViT. In this section, we aim to assess whether this property persists when replacing its patch and embed tokenizer by a SCOTT tokenizer and pretraining within the MIM-JEPA framework. Specifically, we report Top-1 and Top-5 Attentive Probing metrics on Flowers-102 as we scale a SCOTT model along three different axes: (i) pretraining dataset size, (ii) model size, and (iii) pretraining time. While our method is designed to perform well with scarce resources, results in Table \ref{tab:scaling} suggest that not only do SCOTT and MIM-JEPA scale favorably, but they also outperform the standard ViT architecture when computational resources are limited.

\begin{table}[!hbt]
\centering
\caption{Scalability assessment of SCOTT models pretrained on MIM-JEPA.}
\label{tab:scaling}
\begin{tabular}{lcc}
\hline
                                                  & \multicolumn{2}{c}{\textbf{Flowers-102}}                                                    \\
\multirow{-2}{*}{\textbf{Scalability assessment}} & \multicolumn{1}{c}{\textbf{Top-1}}           & \multicolumn{1}{c}{\textbf{Top-5}}           \\ \hline
\rowcolor[HTML]{EFEFEF} 
Pretraining dataset size                          & \multicolumn{1}{c}{\cellcolor[HTML]{EFEFEF}} & \multicolumn{1}{c}{\cellcolor[HTML]{EFEFEF}} \\
1020, i.e. train split (12\%).                     & \multicolumn{1}{c}{74.25}                    & \multicolumn{1}{c}{90.82}                    \\
2040, i.e. train+val (25\%)                        & 80.69                                        & 93.83                                         \\
6149, i.e. test split (75\%)                       & 91.88                                        & 97.70                                         \\
8189, i.e. train+val+test (100\%)                  & 97.15                                        & 99.15                                         \\ \hline
\rowcolor[HTML]{EFEFEF} 
Model size (\# parameters)                         &                                              &                                               \\
SCOTT-3/16 (6.5 M)                                & 93.64                                        & 98.60                                         \\
SCOTT-7/16 (13.6 M)                               & 95.64                                        & 99.07                                         \\
SCOTT-9/16 (17.1 M)                               & 96.50                                        & 99.25                                         \\
SCOTT-12/16 (22.4 M)                              & 97.15                                        & 99.15                                         \\ \hline
\rowcolor[HTML]{EFEFEF} 
Total pretraining time                          &                                              &                                               \\
300 epochs                                               & 97.15                                        & 99.15                                         \\
600 epochs                                               & 97.59                                        & 99.21                                         \\
1200 epochs                                             & 97.73                                        & 99.21                                         \\ \hline
\end{tabular}
\end{table}

\textbf{Scaling data size}. MIM-JEPA pretraining exhibits improved performance when pretrained with larger datasets. This outcome aligns with expectations, as additional data enables the model to learn more general and abstract representations that effectively distinguish between different classes.

\textbf{Scaling model size}. MIM-JEPA pretraining benefits from larger encoder sizes when pretraining on Flowers-102. We increase model sizes by adding more transformer encoder blocks, while keeping the SCOTT tokenizer intact. The predictor network is also kept constant among the different setups.

\textbf{Scaling pre-training time}. A longer MIM-JEPA pretraining time helps the model to produce slightly better image representations.

\section{Code implementation}

To facilitate reproducibility of our work, we release our code at: \url{https://github.com/inescopresearch/scott}





% \end{thebibliography}
\end{document}

%% file: math_commands.tex
\usepackage{amsmath,amsfonts,bm}

\def\eqref#1{equation~\ref{#1}}
\def\1{\bm{1}}

\DeclareMathAlphabet{\mathsfit}{\encodingdefault}{\sfdefault}{m}{sl}
\SetMathAlphabet{\mathsfit}{bold}{\encodingdefault}{\sfdefault}{bx}{n}